\documentclass[review]{elsarticle}

\usepackage{lineno,hyperref}
\usepackage{amsmath}
\usepackage{subcaption}
\usepackage{bm}
\modulolinenumbers[5]
\usepackage[longend]{algorithm2e}
\usepackage{tcolorbox}
\newtcolorbox[auto counter]{Algo}[1]{fonttitle=\bfseries,title={Algorithm \thetcbcounter: #1}}

\usepackage{amssymb}

\journal{Information Fusion}

\begin{document}

\begin{frontmatter}

\title{2D-3D Geometric Fusion Network using Multi-Neighbourhood Graph Convolution for RGB-D Indoor Scene Classification\tnoteref{t1}}

\tnotetext[t1]{Funding: This work was supported by Secretary of Universities and Research of the Generalitat de Catalunya and the European Social Fund via a PhD grant (FI2018) in the framework of project TEC2016-75976-R, financed by the Ministerio de Economía, Industria y Competitividad and the European Regional Development Fund (ERDF).}

\author[1]{Albert Mosella-Montoro\corref{cor1}}
\ead{albert.mosella@upc.edu}

\author[1]{Javier Ruiz-Hidalgo}
\ead{j.ruiz@upc.edu}

\cortext[cor1]{Corresponding author}

\address[1]{Image Processing Group - Deparment of Signal Theory and Communications, Universitat Politècnica de Catalunya, Barcelona, Spain}

\begin{abstract}

Multi-modal fusion has been proved to help enhance the performance of scene classification tasks. This paper presents a 2D-3D Fusion stage that combines 3D Geometric Features with 2D Texture Features obtained by 2D Convolutional Neural Networks. To get a robust 3D Geometric embedding, a network that uses two novel layers is proposed. The first layer, Multi-Neighbourhood Graph Convolution, aims to learn a more robust geometric descriptor of the scene combining two different neighbourhoods: one in the Euclidean space and the other in the Feature space. The second proposed layer, Nearest Voxel Pooling, improves the performance of the well-known Voxel Pooling. Experimental results, using NYU-Depth-V2 and SUN RGB-D datasets, show that the proposed method outperforms the current state-of-the-art in RGB-D indoor scene classification task. 

\end{abstract}


\end{frontmatter}


\section{Introduction}

The scene classification task aims to annotate a sensor capture with a scene class, such as beach, furniture store, bedroom, among others. Due to the rising interest in domotics, surveillance, and robotics applications, the RGB-D indoor scene classification task has received more attention from academia and industry. Despite the advance of the techniques applied to the recognition of object-centric data, these techniques do not have the same performance on indoor scene classification. The main reason is that an indoor scene is formed by a relationship of multiple objects with open-set classes. For instance, recognizing a bed and a chair alone is not enough to classify the scene as a bedroom, because these objects could exist in other scene categories such as a furniture store. Furthermore, there is a data scarcity problem as existing RGB-D datasets are still order-of-magnitude smaller than their respective 2D datasets. Other important challenges that need to be faced in this task are the considerable variation in lights, shapes, and layouts for each class of scene. Most of these challenges have been proved very difficult to solve without the 3D information that is lost in the 2D image capturing. For this reason, the use of stereo-camera configurations, RGB-D sensors, or lidars is recommended. The structure of data captured by these sensors can be organized, like captures done by a Microsoft Kinect sensor, or unorganized, like the information provided by a lidar.

Convolutional Neural Networks (CNNs) are extensively used in computer vision in a wide variety of tasks such as image classification, super-resolution, object detection, and segmentation. Nevertheless, the convolution operation is defined in a lattice structure. That means, data that is not located in a lattice structure can not be processed by CNNs directly as is the case with unorganized 3D point clouds. This limitation can be solved with Geometric Learning, a set of techniques that convert this data to an artificial lattice structure, such as the methods that use voxels to allow the application of 3D CNNs. Another way to handle this kind of data is by using Graph Convolutional Neural Networks (GCNNs). These networks convert a 3D point cloud to a graph and create an artificial lattice structure through the edges of the graph.

This paper proposes a new methodology that fuses the geometric information of a 3D point cloud obtained by the novel Multi-Neighbourhood Graph Convolution network and the 2D texture information obtained by a conventional CNN. The main contributions of this paper can be summarized as:
\begin{itemize}
\item The proposal of the Multi-Neighbourhood Graph Convolution operation, that outperforms previous methods to obtain geometric information. This convolution takes into consideration the neighbours of the center point in Feature and Euclidean spaces.
\item The Nearest Voxel Pooling algorithm, which consists of an improved version of the current Voxel Pooling algorithm~\cite{ecc} that mitigates the noise introduced by sensors.
\item The fusion of 2D-3D and multi-modal features through the proposed 2D-3D Fusion stage. Using geometric proximity allows the network to exploit the benefits of 2D and 3D Networks simultaneously.
\end{itemize}

\section{Related work}

\subsection{Scene classification}

Earlier works of scene classification using RGB information made use of handcrafted features~\cite{sift, hog} to obtain the properties of the scene. Nowadays, with the emergence of deep learning techniques and new datasets, better features can be obtained. Places-CNN~\cite{places365} is a vast dataset of RGB indoor-scene captures that was used to train different standard architectures, such as VGG~\cite{vgg} and ResNet~\cite{resnet}, providing one of the most successful deep feature learning models for RGB data. Instead of finding deep features to describe the scene, \emph{George et al.}~\cite{semantincSceneRecognition} proposed to capture the occurrence statistics of objects in scenes, capturing the informativeness of each detected object for each scene. \emph{Chen et al.} ~\cite{chengraphscenelayout} proposed the Layout Graph Network (LGN), where regions of the scene layout are the nodes and the relations between two independent regions are the edges of the graph. The scene layout is obtained using the proposed method called Prototype-agnostic Scene Layout (PaSL) which constructs a spatial structure without conforming to any prototype. Recently, \emph{Xie et al.}~\cite{xiescenesurvey} published a survey where the existing scene recognition algorithms are reviewed. 

As noted in the introduction, another way to tackle the scene classification problem is by using depth information. \emph{Zhu et al.}~\cite{multimodalfusion2016} proposed to train a two-branch network to learn features from RGB and depth and then fuse these features using an SVM. \emph{Song et al.}~\cite{songlearningeffective} proposed to learn a more effective depth representation using a two-step training approach that directly learns effective depth-specific features using weak supervision via patches. \emph{Li et al.}~\cite{dfnet} proposed a novel discriminative fusion network which can learn correlative and distinctive features of each modality. \emph{Cai et al.}~\cite{rgbdscenecategorizationmultimodal} proposed a multi-modal CNN that captures local structures from the RGB-D scene images and learns a fusion strategy. Similarly, MAPNet~\cite{mapnet} presented two attentive pooling blocks to aggregate semantic cues within and between features modalities. \emph{Song et al.}~\cite{songobjecto-to-object} proposed to use object-to-object relations obtained with detection techniques. More recently, TRecgNet~\cite{trecnet} tackled the RGB-D Scene Recognition problem as a combination of a translation and a classification problem. Their work proposes to train simultaneously a classifier network that classifies the scene and a translation network, that predicts the depth from RGB and the RGB from the depth. Training the network in a multitask manner helps the network learn more generic features that yield an increment in performance. However, these methods use a 2D CNN on depth maps to obtain geometric information that introduces possible errors due to missing local geometric context that the projection to a 2D world can produce. To solve that, in RAGC~\cite{ragc}, authors proposed to extract the geometric information directly from the 3D world. The network exploits the intrinsic geometric context inside a 3D space using as input 3D point clouds obtained from RGB-D captures. However, the nodes of the graph do not have any colour or geometry information.
In this work, the performance of the extraction of the geometric feature is increased using the proposed Multi-neighbourhood Graph Convolution. This convolution fuses two different neighbourhoods, one in the Euclidean space and the other in the Feature space that helps to improve the quality of the extracted features. More details are given in Sec.~\ref{sec:method}.

\subsection{Geometric Learning}

Geometric deep learning consists in a set of emerging techniques attempting to generalize structured deep neural models to non-Euclidean domains or non-structured data such as 3D point clouds. One of the first approaches to process 3D point clouds was the use of Multi-view based techniques~\cite{MultiViewSu3DShapeRecog, SnapNet, SnapNetR, dai3DMV}. These sets of techniques represent a 3D space as a collection of 2D views where the structured deep neural models can be used. However, due to the fact that the 2D view has lost the 3D spatial relation, the geometric information obtained is limited. To work directly in a dense 3D point cloud, different kinds of data structures and network architectures have been proposed, such as voxel grid networks \cite{VoxNet, 3DShapeNets, VolumetricMultiViewCNN, ScanNet, SEGCloud} and octree networks ~\cite{OctreeGeneratingNetworks}. Recently, an strategy that outperforms previous methodologies was presented in DGCNN~\cite{dgcnn}. This strategy consists on representing this data as a graph, where edges are used to create a lattice structure. Two main strategies can be followed to work with these graphs. \emph{Graph Neural Networks}~\cite{ScarselliGraphNeuralNetwork}~\cite{Qui3DGNNSS} where the graph is processed applying a neural network recurrently to every node of the graph, and \emph{Graph Convolutional Networks}~\cite{kipf2017semi}, where a generalization of the discrete convolution is proposed. An improvement of this generalization was proposed by \emph{Wang et al.}~\cite{dgcnn} with the Dynamic Edge Convolution operation. This operation computes each node's feature doing an aggregation over the output of a multi-layer perceptron (MLP) that was applied to the neighbourhood. Following this line of research, \emph{Verma et al.}~\cite{FeaStNetCVPR2018} proposed \emph{FeaStNet}, where the graph-convolution operator consists in establishing correspondences between filter weights and graph neighbourhoods with arbitrary connectivity. More recently, \emph{Mosella-Montoro et al.}~\cite{ragc} presented the Attention Graph Convolution (AGC), which creates an attention weight based on the geometrical attributes of the edges. This convolution outperforms the previous state-of-the-art in the 3D Geometric Scene Classification task. In this work AGC is used as a baseline for the new proposed Multi-Neighbourhood Graph Convolution (MUNEGC).

\subsection{2D-3D Fusion Networks}

The fusion of the features obtained by 2D-3D networks are widely used on multi-view scenarios, where is possible to obtain different 2D RGB images from a dense point cloud. FusionNet~\cite{fusionet} proposed to do a late fusion using the classification scores obtained by the final Fully Connected layers of the 2D and 3D networks. SPLATNet~\cite{splatnet} presented a different approach where the 2D and 3D features representations are mapped onto the same lattice. This mapping is achieved using an improved version of the Bilateral Convolution Layer~\cite{bcl}. More recently, an extension of PointNet++~\cite{pointnet++} for multi-view scenarios with an early fusion strategy was proposed in MVPNet~\cite{mvpnet}. This strategy consists in concatenating the features learned on a 2D-CNN to the geometric point(XYZ) that is used as input of the PointNet++. The main drawback of this approach is that each point of the point cloud used as input on PointNet++ must have a 2D feature associated to it.

These previous methods of 2D-3D fusion made use of multi-view approaches to obtain different 2D RGB images from a single dense point cloud. In the method proposed in this work, the fusion is done using only one 2D RGB image from each point cloud.

\section{Methodology}\label{sec:method}

\subsection{Overview of the framework}\label{sec:overviewframework}

\begin{figure}[htb]
 \centering
 \includegraphics[width=0.8\textwidth]{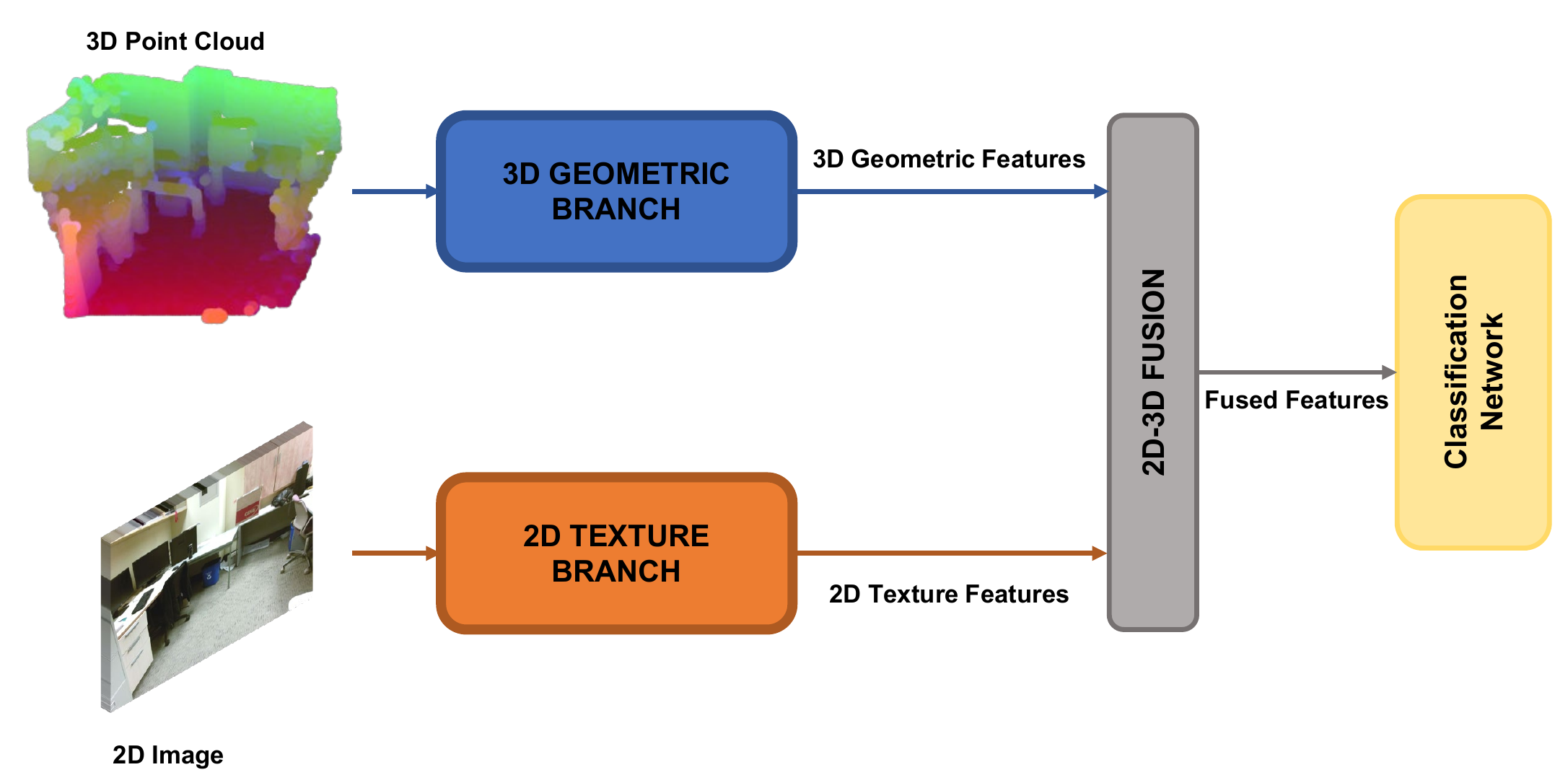}
 \caption{Illustration of the framework of 2D-3D Geometric Fusion network using Multi-Neighbourhood Graph Convolution.}
 \label{fig:3d2doverview}
\end{figure}

The framework proposed is illustrated in Fig.~\ref{fig:3d2doverview}. The framework is composed of two branches: the 3D Geometric branch and the 2D Texture branch. The input of the 3D Geometric branch is a 3D point cloud that can be obtained directly from a lidar sensor or using the depth information and the intrinsic camera parameters of an RGB-D sensor. Each node of the 3D input point cloud encodes the depth information using the HHA encoding~\cite{hha}, that has been proved to obtain better results than directly using the depth channel~\cite{mapnet,trecnet, hha}. HHA encodes the depth into a 0 to 255 range with three channels. Each channel represents horizontal disparity, height above the ground, and the angle with the inferred gravity direction. 
The input of the 2D Texture branch is a 2D RGB image corresponding to the same capture as the capture used on the 3D Geometric branch. After the corresponding branches, the extracted 3D Geometric and 2D Texture features are fused using the 2D-3D Fusion stage, and the result of this stage is used by the Classification network to predict the corresponding scene class.

\begin{figure}[htb]
 \centering
 \includegraphics[width=0.8\textwidth]{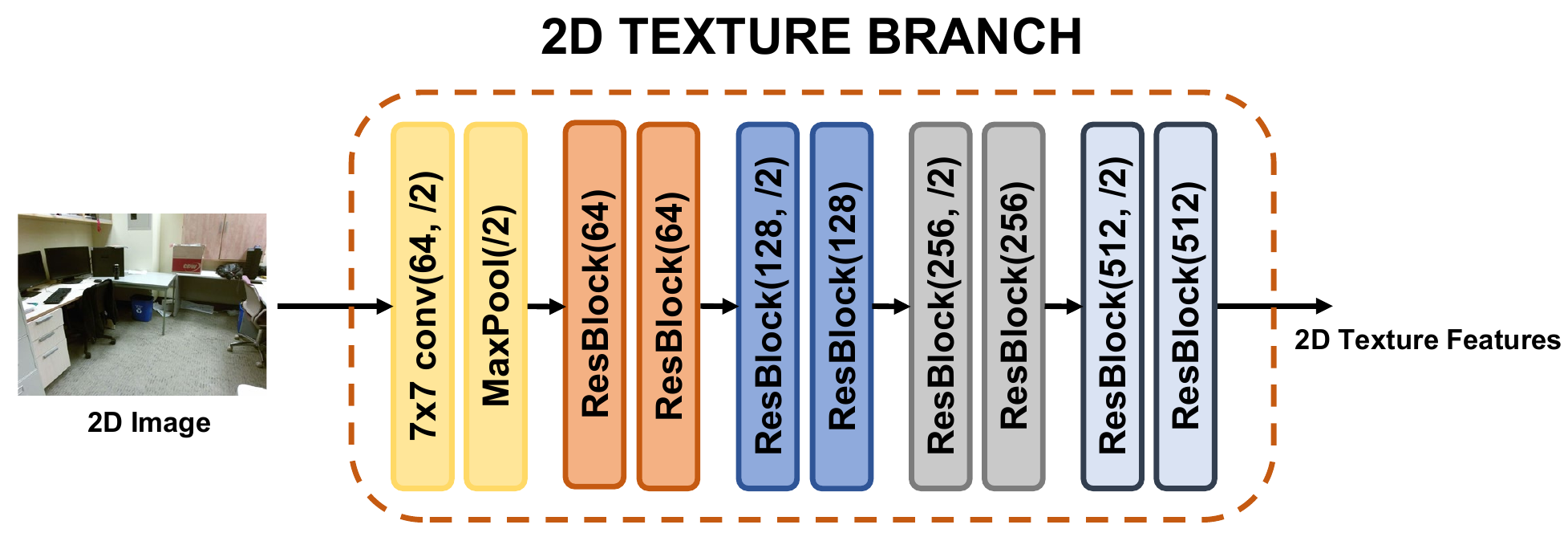}
 \caption{2D Texture branch architecture, where /2 means that the stride has a value of 2 in order to downsample the image by a factor of 2.}
 \label{fig:2dtexturebranch}
\end{figure}

\begin{figure}[htb]
 \centering
 \includegraphics[width=0.4\textwidth]{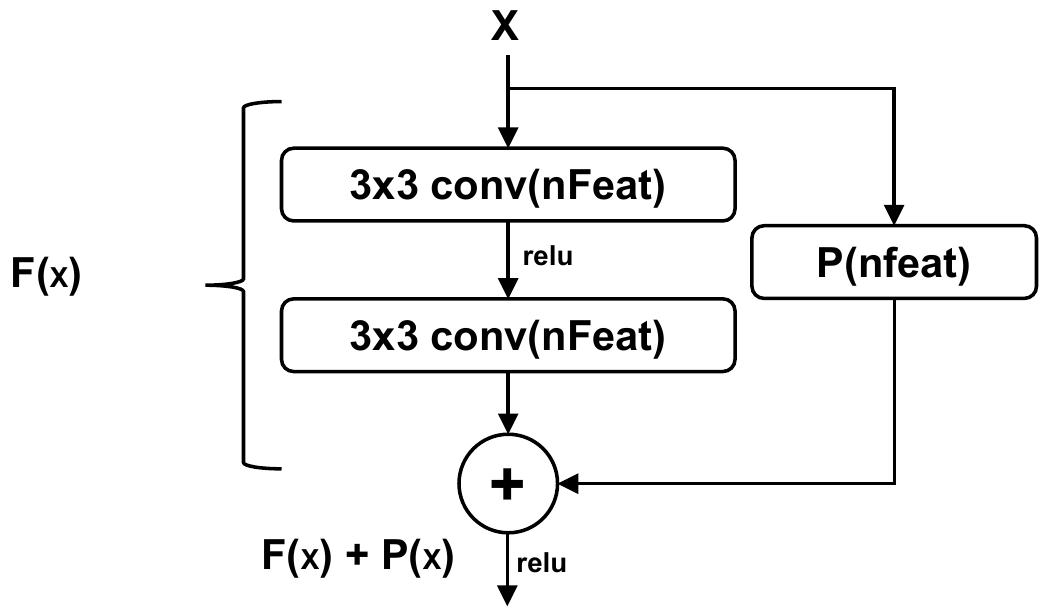}
 \caption{Illustration of a Residual Block.}
 \label{fig:2dresidualblock}
\end{figure}

\textbf{The 2D Texture branch} uses as a backbone the well-known architecture ResNet-18~\cite{resnet}, as depicted in Fig.~\ref{fig:2dtexturebranch}. ResNet-18 is composed of a combination of residual blocks, convolutional layers, and poolings. A Residual Block is a stack of two convolutional layers (F) with a shortcut that contains a projection function (P), as shown in Fig.~\ref{fig:2dresidualblock}. The projection function used by ResNet-18 is a convolutional layer with a kernel size of 1x1 without bias. When the dimension of the input and output features are the same, the projection function is the identity matrix. That kind of block helps to obtain higher accuracy in object-centric classification, reduces the effect of the vanishing gradient problem and accelerates the convergence of the deep networks.

The output of the last Residual block corresponds to the 2D Texture features used for the Fusion stage. This branch aims to exploit the power of already proven CNNs to obtain texture information that will be aggregated to the geometric information obtained by the 3D Geometric branch.

\begin{figure}[htb]
 \centering
 \includegraphics[width=0.8\textwidth]{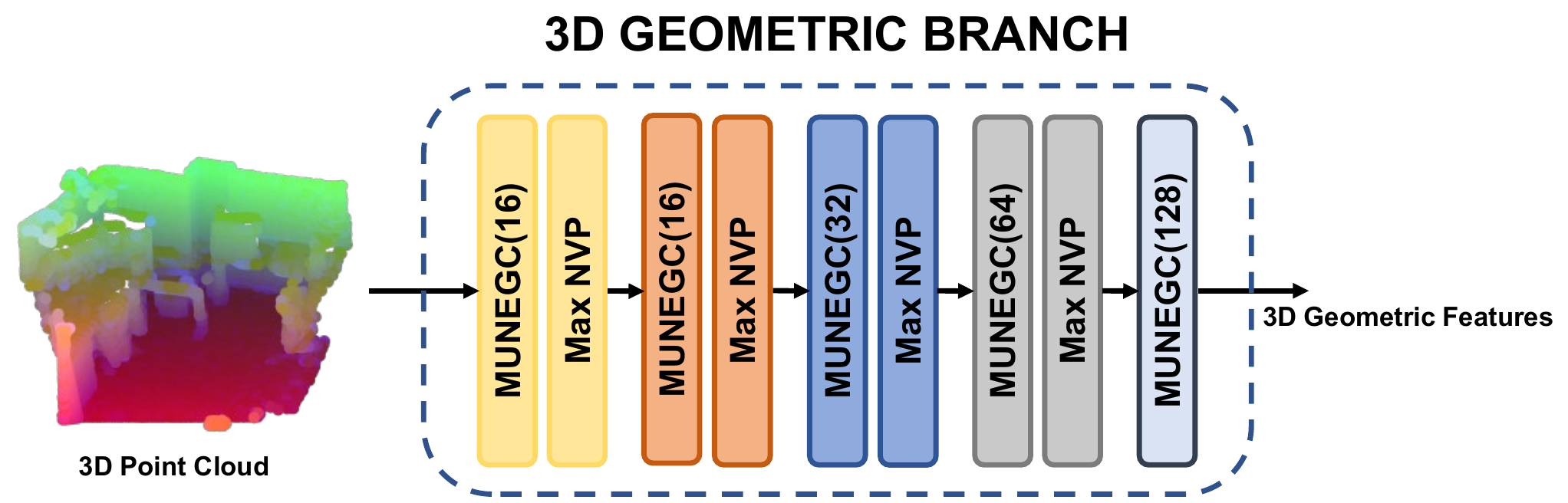}
 \caption{3D Geometric branch architecture.}
 \label{fig:3dgeometricbranch}
\end{figure}

\textbf{The 3D Geometric branch} is composed of two novel layers named Multi-Neighbourhood Graph Convolution (MUNEGC) and Nearest Voxel Pooling(NVP), both layers are explained in detail in Secs.~\ref{sec:munegc} and ~\ref{sec:nvp} respectively. The architecture of the 3D Geometric branch aims to have the same number of pooling stages as ResNet-18. In Fig.~\ref{fig:3dgeometricbranch} the architecture used for the 3D Geometric branch is depicted.

\begin{figure}[htb]
 \centering
 \includegraphics[width=0.4\textwidth]{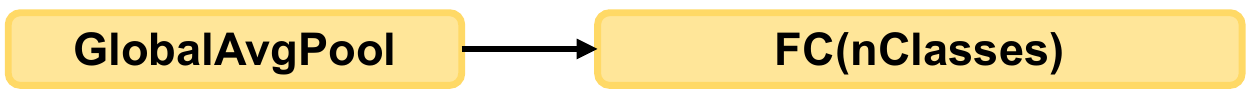}
 \caption{Illustration of the Classification network.}
 \label{fig:classification}
\end{figure}

\textbf{The 2D-3D Fusion stage} takes the features generated by the previously commented 2D and 3D branches and fuses them. Notice that the output resolution and sampling of both branches is different. The reason is that pooling layers of both branches work on different spaces (2D and 3D). As a result, even if 3D point clouds are extracted from RGB-D sensors where the 2D and 3D spaces are organized, the final number of points and their positions are different. The proposed 2D-3D Fusion stage can handle that behaviour and generate a new set of combined features. This stage will be explained in detail in Sec.~\ref{sec:2d-3d fusion}. The new set of features is fed to the Classification network. The classification architecture, as depicted in Fig.~\ref{fig:classification}, is composed of a global average pooling and an $FC(nClasses)$ layer, where $FC$ is a Fully Connected layer and $nClasses$ is the number of scenes that the network should predict. 

\subsection{Multi-Neighbourhood Graph Convolution (MUNEGC)}\label{sec:munegc}

The proposed Multi-Neighbourhood Graph Convolution (MUNEGC) is an extension of the Attention Graph Convolution (AGC)~\cite{ragc}. In order to explain MUNEGC first is necessary to review the bases of the AGC. 

\subsubsection{Review of Attention Graph Convolution(AGC)}\label{subsec:agcreview}

\begin{figure}[htb]
 \centering
 \includegraphics[width=0.7\textwidth]{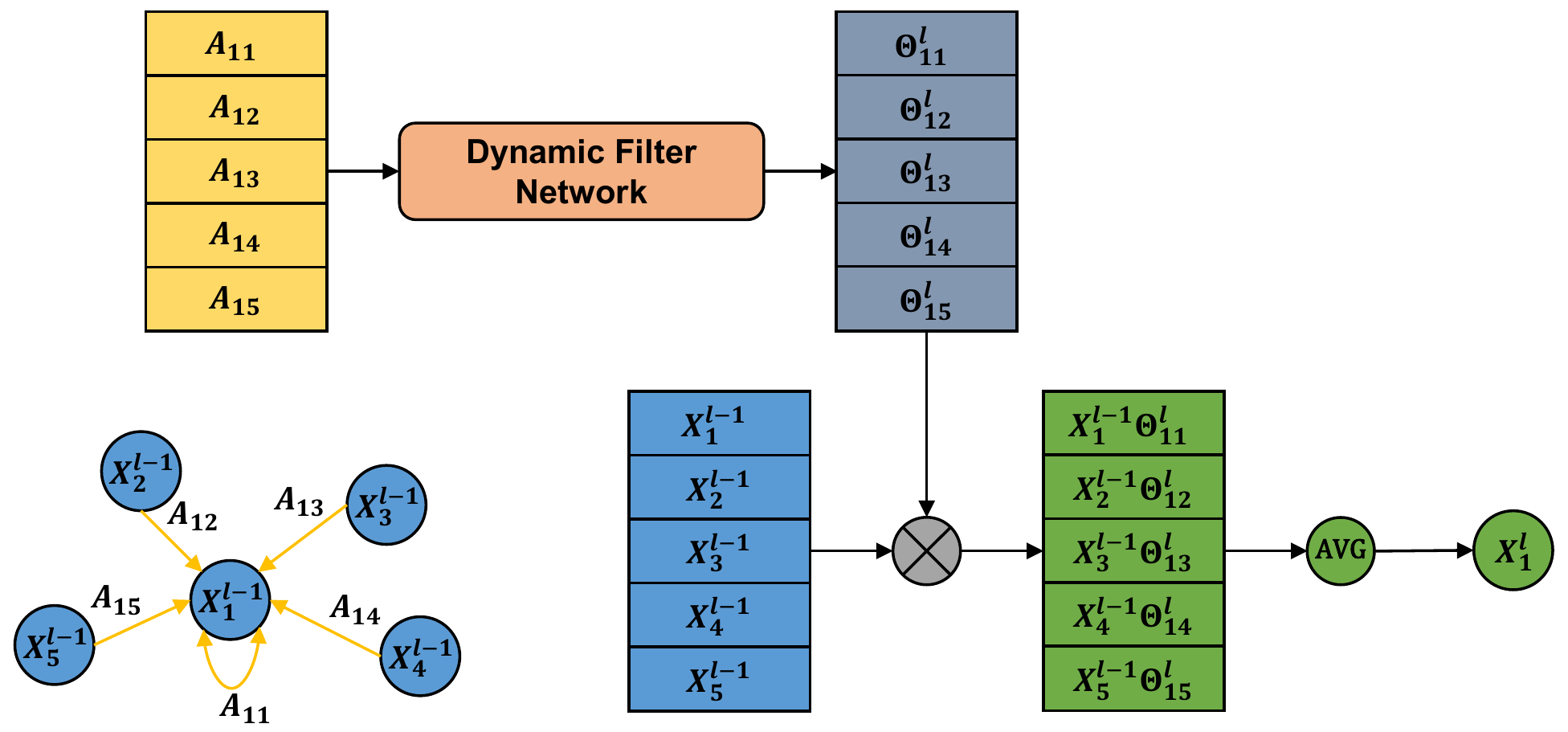}
 \caption{AGC applied to a local neighbourhood. Where $X_i$ is the node feature vector $i$, $A_{ij}$ is the edge's attribute vector of nodes ${ij}$, $\Theta_{ij}$ are the weights generated by the Dynamic Filter Network and $l$ corresponds to a layer index in a feedforward neural network.}
 \label{fig:agc}
\end{figure}

Attention Graph Convolution (AGC)~\cite{ragc} is a graph convolution that performs the convolution over local graph neighbourhoods exploiting the edges and their attributes. These edges are used to create an artificial lattice structure which is needed to apply a convolution. Furthermore, the attributes of the edges are used to estimate the weights of the filter that will be used in each neighbourhood. The generation of weights is based on a \emph{Dynamic Filter Network}~\cite{Dynamicfilter} which can be implemented with any differential architecture. In the case of AGC, the Dynamic Filter network is implemented using $FC(x)$ layers, where $FC$ is a Fully Connected layer and $x$ the number of output features of the layer. The Dynamic Filter Network is in charge of the attention mechanism. It generates weights conditioned by the attributes of the edges. Fig.~\ref{fig:agc} depicts the AGC operation over a node $N_1$ of an input neighbourhood.

In AGC, the attribute of the edge is defined as the positional offset $S_{ij}=p_i-p_j$ between the euclidean position of the nodes $p_i$ and $p_j$ of the edge. These offsets can be represented in euclidean or spherical coordinates. That means, Dynamic Filter Network will pay attention to the nodes depending on their proximity information.

AGC is formalized in Eq.~\eqref{eq:agc} where $X$ is the node feature vector, $N$ the set of neighbours, $W$ represents the Dynamic Filter Network, $A$ represents the edge attributes and $b$ a learnable bias of the layer. Index $i$ indicates the current node to evaluate, $l$ corresponds to a layer index in a feedforward neural network and $j$ the neighbours of the node $i$ in set $N$.

\begin{equation}
 X_i^l = \frac{1}{|N(i)|} \sum_{j\in N(i)} {W}^l(A_{ij}) X_j^{l-1} + b^l
 \label{eq:agc}
\end{equation}

\subsubsection{MUNEGC in detail}

Multi-Neighbourhood Graph Convolution (MUNEGC) is a graph operation that estimates the new feature of each node using the combination of the features obtained in two different neighbourhoods, the Euclidean Neighbourhood and the Feature Neighbourhood.

\begin{figure}[htb]
 \centering
 \includegraphics[width=0.6\textwidth]{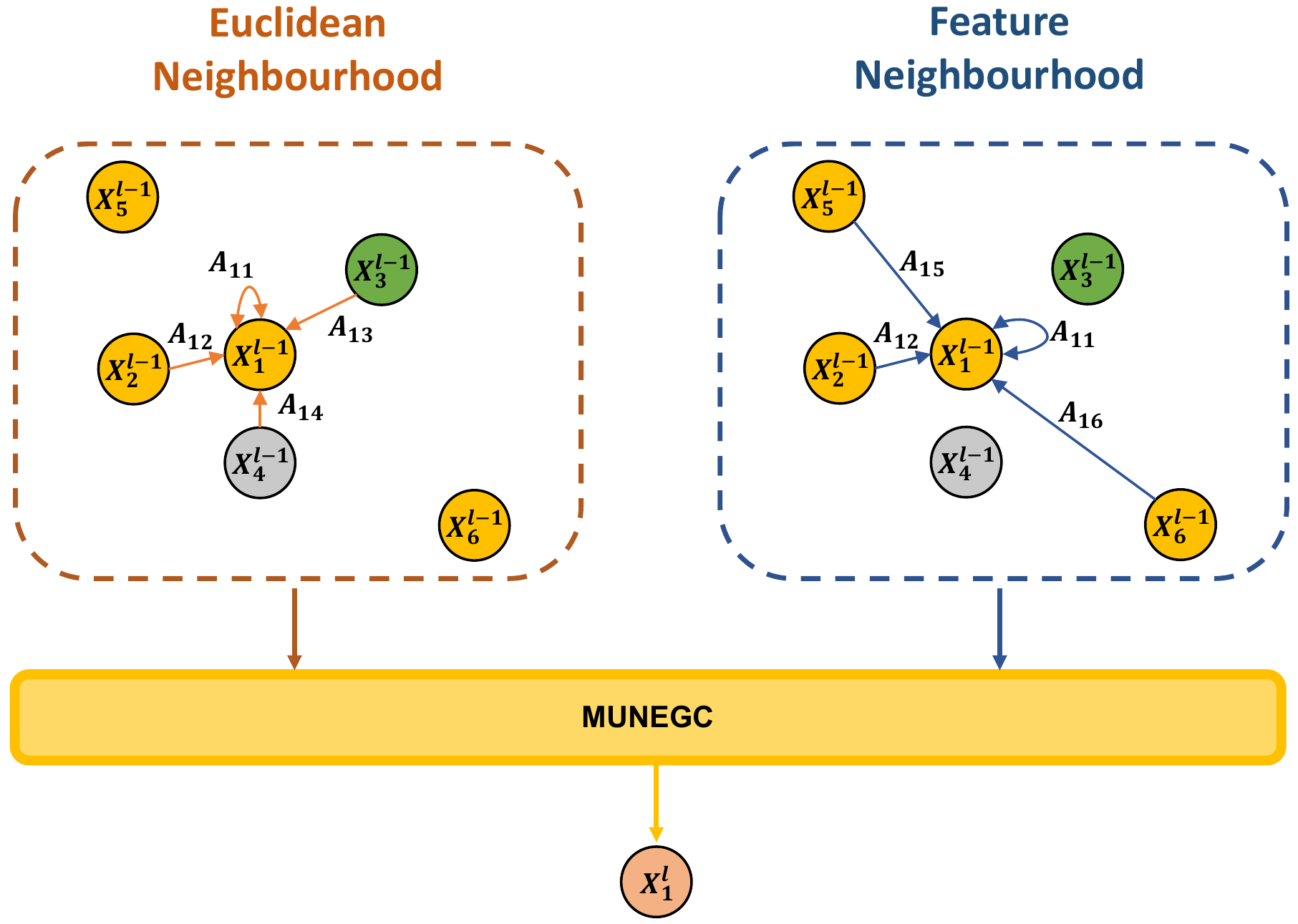}
 \caption{Multi-Neighbourhood Graph Convolution. Where $X_i$ is the node feature vector $i$, $A_{ij}$ is the edge's attribute vector of nodes ${ij}$, and $l$ corresponds to a layer index in a feedforward neural network. Nodes with similar features are coloured with the same colour. Both neighbourhoods are created using a kNN-policy where $k=4$.}
 \label{fig:munegc}
\end{figure}

The neighbourhood creation step consists in connecting each of the nodes to the closest ones with edges. MUNEGC creates two different neighbourhoods in two different spaces, Euclidean and Feature spaces. The Euclidean Neighbourhood uses the position $(x,y,z)$ of the nodes on the Euclidean space to define which nodes are connected. Whereas the Feature Neighbourhood uses the feature vector $X$ of each node to connect the nodes. Therefore, in the Feature Neighbourhood, nodes with similar features (rather than closeness in space) are going to be connected. Both neighbourhoods share the same original nodes and their features. For both neighbourhoods, edges can be generated following a kNN-policy or a Radius-policy. In the case of the Euclidean Neighbourhood, the Radius-policy has a geometric meaning and is intuitive to choose. Whereas in the Feature Neighbourhood, the meaning of this radius is unclear and is not recommended to use due to the complexity of its selection as the Feature space changes at each iteration of the training phase. For that reason, the kNN-policy is chosen in the case of the Feature Neighbourhood. An example of this is depicted in Fig.~\ref{fig:munegc}. Where the same original nodes are used to generate the Euclidean Neighbourhood and the Feature Neighbourhood for the central node $X_i^{l-1}$. For simplicity, both neighbourhoods follow a kNN-policy with $k=4$ (including the self-loop). On the Euclidean Neighbourhood nodes 1, 2, 3 and 4 are selected as neighbours as they are the closest nodes on the 3D space.  Whereas, in the Feature Neighbourhood, nodes 1, 2, 5 and 6 are selected as neighbours as the features of the nodes are similar, represented as nodes with the same colour.

Once both neighbourhoods are defined, attributes for each edge can be computed. MUNEGC proposes to use the combination of the positional offset and feature offset as an attribute of an edge. As in AGC, the positional offset $S_{ij}=p_i-p_j$ is defined as the offset between the euclidean positions $p_i$ and $p_j$ of the nodes $i$ and $j$ interconnected by the edge (positional offset can be represented in euclidean or spherical coordinates). Similarly, the feature offset $K_{ij}=X_i-X_j$ is defined as the offset between the feature vectors $X_i$ and $X_j$ of the nodes interconnected by an edge. Eq.~\ref{eq:edge_feature} formalizes the edge attribute vector proposed in MUNEGC. These attributes are estimated independently for each edge of each neighbourhood. In the end, the attributes of an edge $A_{ij}$ are computed equally for both neighbourhoods.

\begin{equation}
 A_{ij} = \{S_{ij}, K_{ij}\}
 \label{eq:edge_feature}
\end{equation}

As it is done in AGC, MUNEGC uses the attributes of the edges to calculate the weights of the convolution using a Dynamic Filter Network. In order to prevent the network from predicting large weights, this work proposes to apply a $tanh$ activation layer to the predicted weights. 

As a result of having two neighbourhoods, each node will have two possible feature vectors, one for the Euclidean Neighbourhood and another one for the Feature Neighbourhood. By definition the filter size used on both neighbourhoods is the same. This means that, if a MUNEGC of $M$ features is requested, the filters of both neighbourhoods will output $M$ features each one. These features are going to be combined using an aggregation operator. Eqs.~\ref{eq:tanhfilter} and~\ref{eq:munegc} describe the proposed combination by MUNEGC. Where $X$ is the node feature vector, $N$ the set of neighbours, $e$ indicates Euclidean Neighbourhood, $f$ corresponds to the Feature Neighbourhood, index $i$ indicates the current node to evaluate, $l$ corresponds to a layer index in a feedforward neural network, $j$ the neighbours of the node $i$ in set $N$, $W$ represents the Dynamic Filter Network, $A$ represents the edge attributes, $b$ a learnable bias and $Aggr\{ \cdot \}$ represents the maximum or average aggregation operation. The result of the aggregation must be $M$ features.  

\begin{equation}
F(A_{ij})=tanh(W(A_{ij}))
\label{eq:tanhfilter}
\end{equation}

\begin{equation}
\begin{split}
X_i^l = Aggr \biggl\{\frac{1}{|N_e(i)|} \sum_{j\in N_e(i)} F^l_e(A_{ij}) X_j^{l-1} + b_e^l, \\ \frac{1}{|N_f(i)|} \sum_{j\in N_f(i)} F^l_f(A_{ij}) X_j^{l-1} + b_f^l \biggr\}
\end{split}
\label{eq:munegc}
\end{equation}

The use of two different neighbourhoods helps to learn a more robust node feature that considers the characteristics of the regions that have similar properties, and the regions that are close in the 3D space. The use of the positional and feature offset on both neighbourhoods forces the network to learn the influence of a neighbour depending on the similarity of the features and the positional proximity of both points.

\subsection{Nearest Voxel Pooling (NVP)}\label{sec:nvp}

The Nearest Voxel Pooling (NVP) layer is based on the Voxel Pooling (VP) algorithm~\cite{ecc}. The VP algorithm consists in creating voxels of resolution $r_p^l$ over the point cloud and replacing all points inside the voxel with their centroid. The centroid's feature is the average or the maximum of the features of the points inside the voxel. However, VP can introduce some errors when the points inside of two different voxels are closer than their respective voxel's centroid. 

\begin{figure*}[t!]
\begin{subfigure}[t]{.5\textwidth}
 \centering
 \includegraphics[width=.4\linewidth]{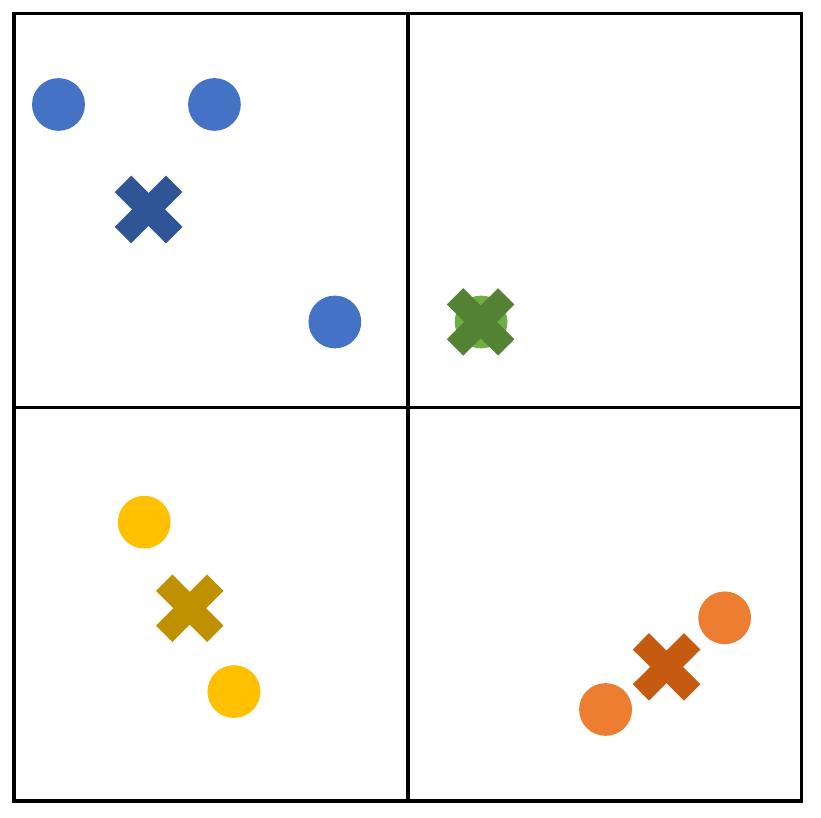} 
 \caption{Voxel pooling example.}
 \label{fig:sub-voxelgrid}
\end{subfigure}%
~
\begin{subfigure}[t]{.5\textwidth}
 \centering
 \includegraphics[width=.4\linewidth]{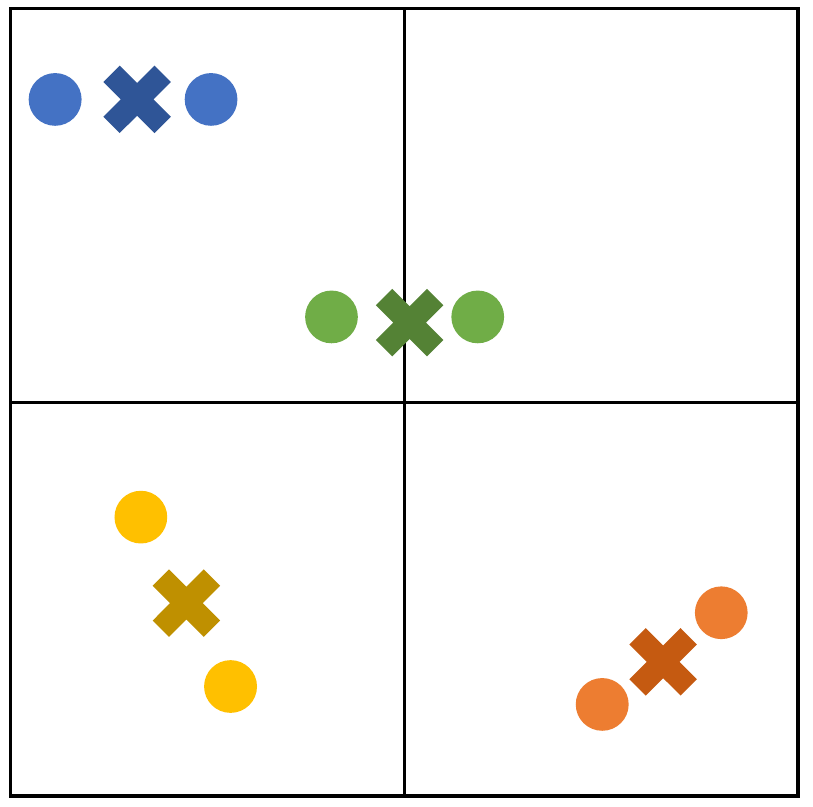} 
 \caption{Nearest voxel pooling example.}
 \label{fig:sub-nvp}
\end{subfigure}
\caption{Comparison of (a) Voxel pooling and (b) Nearest voxel pooling. Crosses represent the new superpoint and the dots the original points. }
\label{fig:fig_vgnvp}
\end{figure*}

The proposed NVP layer reformulates the VP algorithm to solve the issue explained before. In Fig.~\ref{fig:fig_vgnvp}, a comparison of the performance of both algorithms is shown. The algorithm behind the NVP is described in Algo.~\ref{nvp_algorithm} and follows these steps: 

\begin{enumerate}
\item Create voxels of resolution $r_p^l$.
\item Estimate the centroid's position doing the mean of the position of the nodes inside the voxel.
\item For each point of the point cloud, find the closest centroid and group the points that have the same closest centroid.
\item Remove empty voxels and centroids that do not have any point assigned.
\item For each group of points estimate the superpoint's position, doing the mean of the positions of the points inside the group.
\item Superpoint's feature is the average or the maximum of the features of the points that belong to the superpoint's group.
\end{enumerate}

\begin{Algo}{Nearest Voxel Pooling (NVP)}
\begin{algorithm}[H]
Let $r_p^l$ be the radius of the pooling layer($l$) in meters\;
Let $Aggr$ be the aggregation function to aggregate the features\;
Initialize $r_p^l$ and $Aggr$\;
Create voxels $v_i$ of resolution $r_p^l$\;
\ForEach{$v_i$}{
    Compute centroid $c_i$\;
}
\ForEach{point in the PointCloud}{
    Assign point to the closest $c_i$;
}
\ForEach{$c_i$}{
    \uIf{$c_i$ does not have points assigned}{
        Delete $c_i$\;
    }
    \Else{
        position of $c_i$ $\leftarrow$ mean of positions of assigned points\;
        feature of $c_i$ $\leftarrow$ $Aggr$ of features of assigned points\;
  }
}
\label{nvp_algorithm}
\end{algorithm}
\end{Algo}

\subsection{2D-3D Fusion stage}\label{sec:2d-3d fusion}

\begin{figure}[htb]
 \centering
 \includegraphics[width=0.5\textwidth]{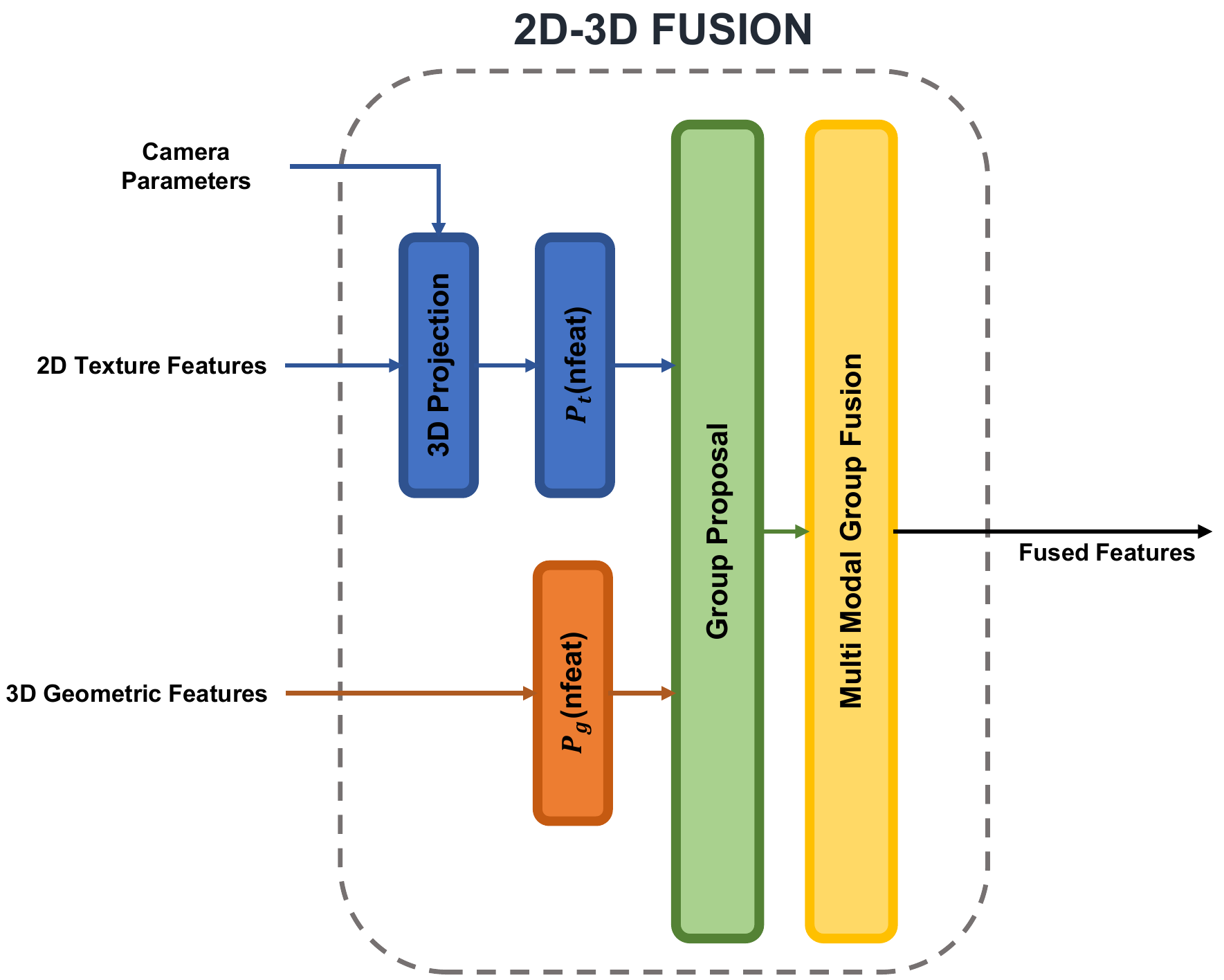}
 \caption{2D-3D fusion stage architecture where $P_g$ is the projection function of the 3D Geometric features and $P_t$ is the projection function of the 2D Texture features.}
 \label{fig:3d2dfusionoverview}
\end{figure}

The 2D-3D Fusion stage is defined to fuse different sets of multi-modal features. This stage will be used in this work to make the fusion of the 3D Geometric and 2D Texture Features. Fig.~\ref{fig:3d2dfusionoverview} shows the architecture proposed for this stage.

The algorithm behind the 2D-3D Fusion stage is described in Algo.~\ref{2D3Dfusion_algorithm} and follows these steps. The 2D Texture Features are projected into the 3D space using the camera parameters. The next step is to apply a Projection function ($P$) in charge of projecting the 3D Geometric and 2D Texture Features to another feature space where both sets of features have the same dimensionality. 

\begin{figure}[htb]
 \centering
 \includegraphics[width=0.3\textwidth]{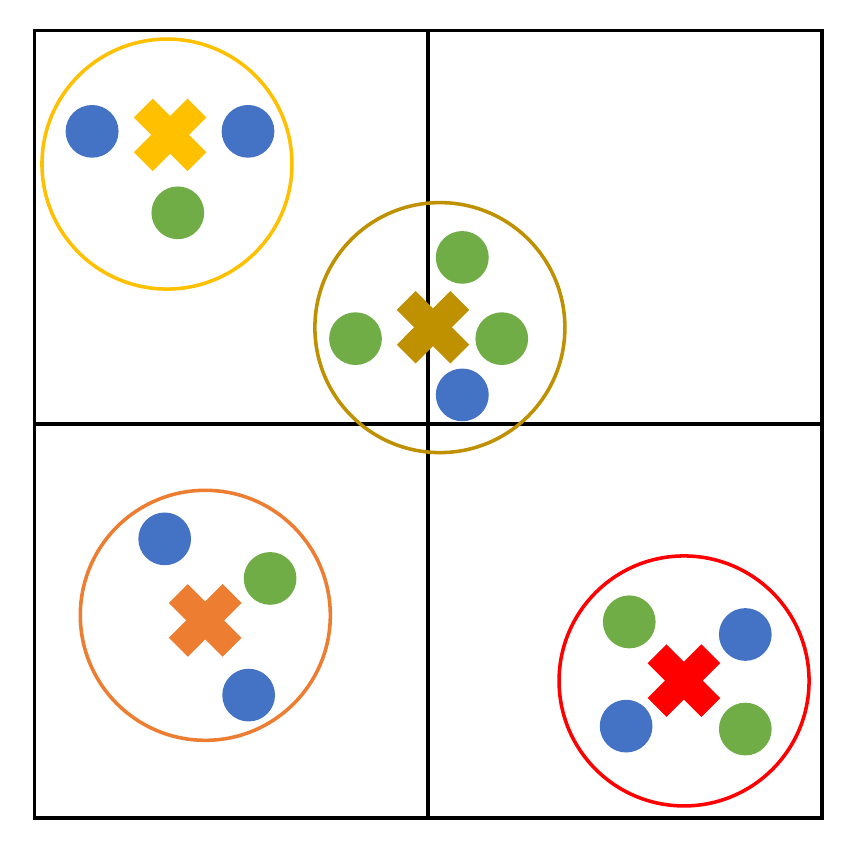}
 \caption{Example of the group's creation for the 2D-3D fusion stage. Each dot colour represents a different kind of feature (blue for 2D Texture and green for 3D Geometric). Each circle represents a different group and the crosses represent the centroid of each group computed by fusing all features of the group.}
 \label{fig:3d2feature_fusion}
\end{figure}

The projected features are fused using a version of the previously explained Nearest Voxel Pooling algorithm, which is used only to create groups of points. Each group of points contains the projected version of the 2D Texture and 3D Geometric features, as shown in Fig.~\ref{fig:3d2feature_fusion}. For each group of points, a superpoint is generated. The position of this superpoint is the average of the positions of the points inside the same group. To estimate the fused feature of each superpoint it is required to follow two steps. First, the average of the same kind of features inside the same group is calculated. Then, the resulting averages are concatenated, generating the fused feature for each superpoint. If there is only one kind of feature in one group, the other positions of the fused vector are filled with a vector of ones. Eq.~\ref{eq:superpointfeature} formalizes the described fusion,  where $X$ is a feature vector, $P$ is the projection function and $N$ is the set of points inside the same group. Index $i$ indicates the current group to evaluate, $g$ indicates that it belongs to the 3D geometric feature subset and $t$ indicates that it belongs to the 2D Texture feature subset. 

\begin{equation}
X_i = Concat \left\{\frac{1}{|N_g(i))|} \sum_{(X\in N_g(i))}P_g(X), \frac{1}{|N_t(i))|} \sum_{X\in N_t(i))}P_t(X) \right\}
\label{eq:superpointfeature}
\end{equation}

Notice that the motivation of the Projection function ($P$) is that each kind of input features could have different dimensionality. To solve that, $P$ projects both input features to another feature space where both sets of features have the same dimensionality. Function $P_g$ is used to project 3D Geometric features and $P_t$ for projecting 2D Texture features. Both are defined as a convolutional layer with a kernel size of 1x1 without bias. 

\begin{Algo}{2D-3D Fusion}
\begin{algorithm}[H]
Let $r$ be the radius used to fuse the features in meters\;
Let $P_t$ be the Projection function used on 2D Texture Features($TF$)\;
Let $P_g$ be the Projection function used on 3D Geometric Features($GF$)\;

Project 2D Texture Features to the 3D space\;
Apply $P_t$ to the 2D Texture features\;
Apply $P_g$ to the 3D Geometric features\;

Create voxels $v_i$ of resolution $r$\;

\ForEach{$v_i$}{
    Compute centroid $c_i$\;
}
\ForEach{point in the 3D space}{
    Assign point to the closest $c_i$;
}
\ForEach{$c_i$}{
    \uIf{$c_i$ does not have points assigned}{
        Delete $c_i$\;
    }
    \Else{
        $TC$ $\leftarrow$ mean of $TF$ of assigned points\;
        $GC$ $\leftarrow$ mean of $GF$ of assigned points\;
        feature of $c_i$ $\leftarrow$ $Concat$ of  $TC$ and  $GC$\;
        position of $c_i$ $\leftarrow$ mean of positions of assigned points\;
  }
}
\label{2D3Dfusion_algorithm}
\end{algorithm}
\end{Algo}

\section{Experiments}
\subsection{Datasets}\label{sec:datasets}

\textbf{The SUN RGB-D dataset}~\cite{SUNRGBD} includes 10335 RGB-D captures. The dataset was captured from different RGB-D sensors including Asus Xtion, RealSense, Kinect v1 and Kinect v2. Following the settings proposed by the authors, classes with less than 80 samples are discarded. In the end, 9504 captures remain with 19 different classes. These captures were divided in 4845 for training and 4659 for testing using the official split.

\textbf{The NYU-Depth-V2 dataset (NYUV2)}\cite{NYUV2} contains 1449 RGB-D captures with 27 classes. Following the standard configuration, the categories are grouped into $10$, including $9$ most common categories and the \emph{Other} category representing the rest. The standard split is followed, where 795 captures are used in the train split and 654 for testing.

\subsection{Training details}\label{sec:traindetails}

The implementation of the proposed framework is based on Pytorch~\cite{pytorch} and Pytorch Geometric~\cite{pygeometric} and can be found at: \url{https://imatge-upc.github.io/munegc/}. Due to GPU memory constraints, each branch of the network is trained in an isolated manner adding an independent classification network for each branch. The Classification network is initialized randomly where biases are initialized as $b = -log((1-\pi)/\pi)$ where $\pi = 1/C$ and $C$ the number of classes. This initialization aims to avoid the possible training instability that bias $b=0$ could cause at the beginning of the training as explained by \emph{Cui et al.}~\cite{nsamples}. 

\subsubsection{Input data}\label{sec:inputdata}

Both datasets used are composed by RGB-D captures. The RGB part of the capture is used as input by the 2D Texture branch. In both datasets the RGB image has been cropped using a center-crop technique obtaining a new resolution of $560 \times 420$. The 3D Geometric branch uses as input the 3D projected version of the depth capture. In particular, the depth capture is encoded using the HHA encoding~\cite{hha}, downsampled by a factor of $x8$ and projected to the 3D space using the parameters of the camera. Let $[x, y, z]$ be the 3D coordinates in the camera coordinate system and $[u, v]$ the coordinates in the image. The focal length of the camera is represented by $[f_x, f_y]$ and the principal point is represented as $[c_x, c_y]$. To project the depth capture to a 3D space the Pinhole camera model is used, Eq.~\eqref{eq:pinhole_projection} formalizes the projection used. 

\begin{equation} \label{eq:pinhole_projection}
\left.
\begin{aligned}
    z &= depth channel\\
    x &= \frac{(u - c_x) \cdot z}{f_x}\\
    y &= \frac{(v - c_y) \cdot z}{f_y}
\end{aligned}
\right\}
\end{equation} 

Both datasets are characterized by having an unbalanced number of images for each category. This work uses the well known weighted cross-entropy (WCE) loss to handle the imbalance issue during training in all branches. The re-scaling strategy used is defined in Eq.~\ref{eq:bce}. Where $x$ is the output vector of the network, $y$ is the ground truth label, and $w$ is the weight vector that contains a different weight for each class. The $w(y)$ is computed using the inverse class frequency, $f(y)$, as described in Eq.~\ref{eq:weighticf}.

\begin{equation}
WCE(x, y) = w[y] \left(-x[y] + \log\left(\sum_j e^{(x[j])}\right)\right)
\label{eq:bce}
\end{equation}

\begin{equation}
w(y) = 1/f(y)
\label{eq:weighticf}
\end{equation}

Furthermore, to make the total loss in the same scale when the weight is applied, $w(y)$ is normalized so that $\sum_{t=1}^{C}w(y) = C$ where $C$ is the total number of classes.

\subsubsection{2D Texture branch training details}

The 2D Texture branch, as explained in Sec.~\ref{sec:method}, uses ResNet-18 as backbone as it is done in recent state of the art~\cite{trecnet}. Similar to previous works~\cite{trecnet, mapnet}, weights are initialized using a pre-trained version of the network on Places dataset~\cite{places365} in the SUN RGB-D dataset. For the smaller NYUV2 dataset, ResNet-18 is initialized using the weights obtained on the training done in the SUN RGB-D dataset. In both datasets, the network is trained during 100 epochs with a batch size of 16. The optimizer used for this training is SGD with momentum. The learning rate used is $1\times10^{-3}$ with a momentum of $0.9$ and a weight decay of $1\times10^{-4}$. A random horizontal flip is applied during training.

\subsubsection{3D Geometric branch training details}

The 3D Geometric branch is initialized randomly for the SUN RGB-D dataset as there is no bigger RGB-D dataset to perform a pre-training of the network. In the case of the NYUV2 dataset, as done in the 2D Texture branch, weights obtained on SUN RGB-D are used to initialize the branch. The motivation for this is to demonstrate the ability of the 3D Geometric branch to learn generalized representations that can be used on other datasets as proved in Sec.~\ref{resultsnyuv2}. In both datasets, the network is trained during 200 epochs with a batch size of 32. The optimizer used for this training is the Rectified Adam (RADAM)~\cite{radam}, an improved version of ADAM that rectifies the variance of the adaptive learning rate. The learning rate used is $1\times10^{-3}$, betas $(0.9, 0.999)$ and a weight decay of $1\times10^{-4}$. A dropout layer is added before the Fully Connected layer of the Classification network with a probability $p=0.2$ to be zeroed. The radius chosen for the pooling layers can be seen in Table~\ref{tab:radius_pooling}. The Dynamic Filter Network configuration chosen for all MUNEGC layers is: $FC(128)- FC(d_l \cdot d_{l-1})$ where $d_l$ is the number of output features of layer $l$. The average aggregation is used in MUNEGC layers.  

The input data used in this network is a 3D Point Cloud that contains for each node an HHA feature, the procedure to obtain this 3D Point Cloud has been described in Sec.~\ref{sec:inputdata}. The policy used by MUNEGC to obtain both versions of the neighbourhoods is the kNN-policy with $k=9$. The edges of the graph have as attributes the positional offset in spherical coordinates and the feature offset as described in Sec.~\ref{sec:munegc}.

Finally, the following techniques of online data augmentation are applied: 1) Rotation over the vertical axis randomly between $(0, 2\pi)$. 2) Mirroring over horizontal axis randomly with a probability of $0.5$. 3) Random removal of points in the input 3D point cloud with a probability of 0.2. 4) A novel 3D random crop proposed in this work. This technique consists in finding a random centroid, then, for each axis, a random value between the maximum and minimum is chosen. A factor $f$ is defined to specify the desired number of points inside the crop. The values of $f$ are in the range of $0<f<1$. The desired number of points ($dn$) is defined as $dn = npoints \times f$, where $npoints$ is the number of points of the original point cloud.  Finally, a radius that accomplishes the following condition $npoints\_inside < dn$ is found, where $npoints\_inside$ indicates the number of points inside the proposed radius. The crop is made up of the points inside the sphere defined by the radius found. In this work, the $f$ is randomly chosen inside the range $[0.875, 1]$.

\begin{table}[ht]
\begin{center}
\begin{tabular}{c c}
\hline
Pooling Layer & Radius(meters)\\ 
\hline \hline
PNV 1 & 0.05\\
PNV 2 & 0.08\\
PNV 3 & 0.12\\
PNV 3 & 0.24\\
\hline
\end{tabular}
\end{center}
\caption{Pooling radius configuration of 3D Geometric Branch.}
\label{tab:radius_pooling}
\end{table}

\subsubsection{2D-3D Fusion and Classification stage training details}
The 2D-3D Fusion stage and the Classification network are considered the last branch, and both networks are trained together. In both datasets, weights are initialized randomly. The input of this branch is the camera parameters and the output features of both previous branches, without its corresponding classification networks as it can be seen in Fig.~\ref{fig:3d2doverview}. The network is trained during 20 epochs with a batch size of 32. Rectified Adam (RADAM)~\cite{radam} is used for the training with a learning rate of $1\times10^{-3}$, betas $(0.9, 0.999)$ and a weight decay of $1\times10^{-4}$. A dropout layer is added before the Fully Connected of the classification network with a probability $p=0.5$ to be zeroed. The radius used to fuse the features in the 2D-3D Fusion stage is $r=0.24$.

\subsection{Results on SUN RGB-D dataset}

\subsubsection{Comparison with state-of-the-art methods}

In this section, the final results of the proposed method are compared with the most recent state-of-the-art in Indoor Scene Classification. Previous methods use a pre-trained 2D-CNN to obtain geometric and texture features. Multi-modal fusion~\cite{multimodalfusion2016} used two independent CNNs to obtain features from RGB and depth data, the fusion is done using an SVM. DF$^2$Net~\cite{dfnet} made use of a triplet loss to encourage the network to learn discriminative and correlative features to do a better fusion. Effective RGB-D representations~\cite{songlearningeffective} learned effective depth-specific features using weak supervision via patches. MAPNet~\cite{mapnet} improved the fusion stage, adding two attentive pooling blocks to aggregate semantic cues within and between features modalities. TRecNet~\cite{trecnet} proposed to use a combination of a translate and classification problem, that predicts the depth from RGB and the RGB from the depth. This approach allowed TRecNet to obtain more generic features and extra data that can be used in training. The method proposed in this paper tries to improve the performance using a completely different approach. First, the geometric features are obtained in a 3D space using the proposed graph convolution MUNEGC and the pooling layer NVP, that allows the network to exploit the intrinsic geometric context inside a 3D space. The fusion strategy is tackled by the novel 2D-3D Fusion stage that, using geometric proximity, allows the network to exploit the benefits of 2D and 3D Networks simultaneously. With these contributions, the proposed method improves previous state-of-the-art methods with an increment of $1.9\%$ in the mean accuracy, as shown in Table ~\ref{tab:sun_results}.

\begin{table}[ht]
\begin{center}
\begin{tabular}{c c c c}
\hline
& & Mean Acc(\%)\\
Method & RGB & Geometric & Fusion\\ 
\hline \hline
Multi-modal fusion~\cite{multimodalfusion2016} & 40.4 & 36.5 & 41.5\\
Effective RGB-D representations~\cite{songlearningeffective} & 44.6 & 42.7 & 53.8\\
DF$^2$Net~\cite{dfnet} & 46.3 & 39.2 & 54.6\\
MAPNet~\cite{mapnet} & - & - & 56.2\\
TRecNet~\cite{trecnet} & 50.6 & 47.9 & 56.7\\
\textbf{Ours} & 56.4 & 44.1 & \textbf{58.6}\\
\hline
\end{tabular}
\end{center}
\caption{Performance comparison with state-of-the-art methods on SUN RGB-D Dataset.}
\label{tab:sun_results}
\end{table}

One of the disadvantages of the proposed method is the lack of large RGB-D datasets to do a proper pre-training of the 3D Geometric branch. Table~\ref{tab:sun_results_geometric} compares the mean accuracy using only geometric features in TRecNet, that is the best method on the current-state-of-the-art and the 3D geometric features of the proposed framework. It shows that using pre-trained features helps TRecNet improve up to $5.4\%$ the mean accuracy. However, the proposed 3D Geometric branch exceeds the mean accuracy of TRecNet when initialized randomly. 

\begin{table}[ht]
\begin{center}
\begin{tabular}{c c c}
\hline
Method & Initialization & Mean Acc.(\%)\\ 
\hline \hline
\textbf{TRecNet~\cite{trecnet}} & \textbf{Places} & \textbf{47.6}\\
TRecNet~\cite{trecnet} & Random & 42.2\\
\underline{Ours}& \underline{Random} & \underline{44.1}\\
\hline
\end{tabular}
\end{center}
\caption{Performance comparison of the 3D Geometric branch with state-of-the-art methods on SUN RGB-D Dataset.}
\label{tab:sun_results_geometric}
\end{table}

\subsubsection{Study of different strategies to generate the neighbourhood and the attributes of the edges}

In this section, different strategies to generate the neighbourhood and their respective edge attributes will be studied. All parameters and configurations explained in Sec.~\ref{sec:traindetails} are fixed. In order to generate the neighbourhoods, two different policies can be used: kNN and Radius. However, the Radius-policy can not be applied in the Feature Neighbourhood as features are still being defined during the training phase, which complicates the choice of a radius. For this reason, the study of the two main policies will be done in the Euclidean Neighbourhood. A radius needs to be chosen for each MUNEGC layer. The best radius found where:  $[0.05m, 0.08m, 0.12m, 0.24m, 0.48m]$ for each corresponding MUNEGC layer. As it can be observed in Table~\ref{tab:ablation_knn_vs_radius}, kNN-policy surpasses the mean accuracy obtained with Radius-policy in this scenario.

\begin{table}[ht]
\begin{center}
\begin{tabular}{c c}
\hline
Method & Mean Acc.(\%)\\ 
\hline \hline
\textbf{kNN-policy} & \textbf{44.1}\\
Radius-policy & 42.44\\
\hline
\end{tabular}
\end{center}
\caption{Analysis of kNN and radius as edge generation policy in Euclidean Neighbourhood.}
\label{tab:ablation_knn_vs_radius}
\end{table}

Once the neighbourhood is defined, an attribute for each edge should be assigned. In Table~\ref{tab:ablation_edge_features} an analysis of the different edge's attributes on each kind of neighbourhood can be seen. The best configuration is the use of Spherical offset and Feature offset on both neighbourhoods. As can be seen, both offsets are required in both neighbourhoods, as explained in Sec.~\ref{sec:munegc}. 

\begin{table}[ht]
\begin{center}
\small
\begin{tabular}{c c c}
\hline
Euclidean attributes & Feature attributes & Mean Acc. (\%)\\ 
\hline \hline
\textbf{Spherical $+$ Feature offsets} & \textbf{Spherical $+$ Feature offsets} & \textbf{44.1}\\
Cartesian $+$ Feature offsets & Cartesian $+$ Feature offsets & 42.27\\
Spherical $+$ L2 offsets & Spherical $+$ L2 offsets & 40.45\\
Spherical offset  &  Feature offset & 40.24\\
\hline
\end{tabular}
\end{center}
\caption{Analysis of the effectiveness of different edge attributes on each kind of neighbourhood. L2 offset is the L2 distance between the feature vector of two neighbours.}
\label{tab:ablation_edge_features}
\end{table}

\subsubsection{Analysis of the MUNEGC design}

In this section, the MUNEGC design will be analyzed. As it is explained in Sec.~\ref{sec:munegc}, apart from the Multi-neighborhood convolution, MUNEGC proposes two extensions to the vanilla AGC~\cite{ragc}. The first one is to add the node feature offset as an attribute of the edge. The second one is to create a mechanism to prevent the prediction of large weights by the Dynamic Filter Network that can cause unstable learning. A $tanh$ is added as an activation layer at the end of the network. In Table~\ref{tab:ablation_agcvsmunegc}, the influence of each one of these extensions can be observed.

\begin{table}[ht]
\begin{center}
\begin{tabular}{c c c c c}
\hline
layer & Spherical offset & Feature offset & tanh & Mean Acc.(\%)\\ 
\hline \hline
AGC & Yes & No & No & 35.7\\
AGC & Yes & Yes & No & 41.53\\
AGC & Yes & Yes & Yes & 42.37\\
\textbf{MUNEGC} & \textbf{Yes} & \textbf{Yes} & \textbf{Yes} & \textbf{44.1}\\
\hline
\end{tabular}
\end{center}
\caption{Analysis of the performance of each improvement done in MUNEGC.}
\label{tab:ablation_agcvsmunegc}
\end{table}

Furthermore, in Table~\ref{tab:ablation_munegc_aggr}, the influence of the aggregation method of both neighbourhoods in MUNEGC can be seen. The average aggregation shows a better performance than the maximum aggregation.

\begin{table}[ht]
\begin{center}
\begin{tabular}{c c}
\hline
Method & Mean Accuracy(\%)\\ 
\hline \hline
\textbf{Average} & \textbf{44.1}\\
Maximum & 40.7\\
\hline
\end{tabular}
\end{center}
\caption{Comparison between maximum and average aggregation in MUNEGC.}
\label{tab:ablation_munegc_aggr}
\end{table}

\subsubsection{Analysis of the Nearest Voxel Pooling}

The Nearest Voxel Pooling (NVP) is an improved version of the Voxel Pooling (VP) that solves the drawback of VP when the points inside of two different voxels are closer than their respective voxel’s centroid. NVP layers are replaced with VP layers to analyze the performance of the proposed NVP in the 3D Geometric branch. Table~\ref{tab:ablation_nvp} shows that the NVP achieves better results than VP.

\begin{table}[ht]
\begin{center}
\begin{tabular}{c c}
\hline
Method & Mean Acc.(\%)\\ 
\hline \hline
\textbf{NVP} & \textbf{44.1}\\
VP & 42.5\\
\hline
\end{tabular}
\end{center}
\caption{Comparison between Nearest Voxel Pooling (NVP) and Voxel Pooling (VP) algorithms.}
\label{tab:ablation_nvp}
\end{table}

\subsubsection{Analysis of the Fusion Stage}

In this section, the proposed 2D-3D Fusion stage is analyzed. This fusion is performed using geometric proximity, where features extracted from the 2D Texture and 3D Geometric branches are fused in the 3D domain. To prove the validity of the proposed method, a comparison with a late fusion stage is performed. The late fusion is inspired by the one used by FuseNet~\cite{fusionet}. It concatenates the features obtained from each branch after a global average pooling. Therefore, only a concatenation is used without taking into account any geometric proximity. 

In addition, the influence of the group algorithm is studied. Both NVP and VP algorithms has been compared to create the group of points to fuse. As it can be seen in Table~\ref{tab:ablation_fusion stage}, the proposed 2D-3D Fusion stage with NVP outperforms the results obtained using the VP algorithm and the late fusion stage.

\begin{table}[ht]
\begin{center}
\begin{tabular}{c c}
\hline
Method &  Mean Acc.(\%)\\ 
\hline \hline
\textbf{Proposed with NVP} & \textbf{58.6}\\
Proposed with VP & 57.8\\
Late & 57.19\\

\hline
\end{tabular}
\end{center}
\caption{Analysis of the validity of the proposed 2D-3D Fusion stage.}
\label{tab:ablation_fusion stage}
\end{table}

\subsection{Results on NYU-Depth-V2 dataset}\label{resultsnyuv2}

The proposed frameworks are also evaluated on the NYUV2 dataset and compared with the state-of-the-art. Unlike the experiments done in SUN RGB-D dataset, the 3D Geometric branch can be pre-trained using the weights obtained from the training on SUN RGB-D. As it can be seen in Table~\ref{tab:nyu_results}, the proposed method overcomes the state-of-the-art by $6\%$ of the mean accuracy. Experiments on NYUV2 reveal that the proposed 3D Geometric branch composed by MUNEGC and NVP has the ability to learn generalized representations that can be used on other datasets, making it possible to apply transfer learning techniques as conventional 2D-CNNs.

\begin{table}[ht]
\begin{center}
\begin{tabular}{c c c c}
\hline
& & Mean Acc(\%)\\
Method & RGB & Geometric & Fusion\\ 
\hline \hline
Effective RGB-D representations~\cite{songlearningeffective} & 53.4 & 56.4 & 67.5\\
DF$^2$Net~\cite{dfnet} & 61.1 & 54.8 & 65.4\\
MAPNet~\cite{mapnet} & - & - & 67.7\\
TRecNet~\cite{trecnet} & 64.8 & 57.7 & 69.2\\
\textbf{Ours} & 67.8 & 59.2 & \textbf{75.1}\\
\hline
\end{tabular}
\end{center}
\caption{Performance comparison with state-of-the-art methods on NYU-Depth-V2 Dataset.}
\label{tab:nyu_results}
\end{table}

The comparison in performance between the proposed 3D Geometric branch and TRecNet, which is the best method on the current state-of-the-art, can be seen in Table~\ref{tab:nyu_results_geometric}. As it can be seen, when the proposed method is initialized randomly, it has better accuracy than TRecNet when this one is initialized with Places. Furthermore, the proposed method outperforms TRecNet when both are initialized with the same dataset.

\begin{table}[ht]
\begin{center}
\begin{tabular}{c c c}
\hline
Method & Initialization & Mean Acc.(\%)\\ 
\hline \hline
TRecNet~\cite{trecnet} & Places & 55.2\\
TRecNet~\cite{trecnet} & SUN RGB-D & 57.7\\
\underline{Ours} & \underline{Random} & \underline{57.2}\\
\textbf{Ours} & \textbf{SUN RGB-D} & \textbf{59.2}\\
\hline
\end{tabular}
\end{center}
\caption{Performance comparison of the 3D Geometric branch with state-of-the-art methods on NYU-Depth-V2 dataset.}
\label{tab:nyu_results_geometric}
\end{table}

\section{Conclusions}

This paper proposes a 2D-3D Geometric Fusion Network that exploits the intrinsic geometric information of the 3D-space to obtain geometric features and improves the fusion with the texture features. The geometric features are obtained by the 3D Geometric branch that is composed by Multi-Neighbourhood Graph Convolution (MUNEGC) and Nearest Voxel Pooling (NVP) layers. The 2D Texture features are obtained by a standard 2-CNN as ResNet-18. The 2D-3D Fusion stage exploits 3D geometric proximity to fuse both the 3D Geometric features and the 2D Texture features. As experiments demonstrate on SUN RGB-D and NYU-Depth-V2 dataset, the proposed method outperforms state-of-the-art results validating the effectiveness of the proposed layers and stages. One direction of the future work is to explore the possibility of transfering the knowledge obtained by a pre-trained 2D-CNN to the proposed MUNEGC network using translation.


\bibliography{bibtex}

\begin{thebibliography}{10}
\expandafter\ifx\csname url\endcsname\relax
  \def\url#1{\texttt{#1}}\fi
\expandafter\ifx\csname urlprefix\endcsname\relax\def\urlprefix{URL }\fi
\expandafter\ifx\csname href\endcsname\relax
  \def\href#1#2{#2} \def\path#1{#1}\fi

\bibitem{ecc}
M.~{Simonovsky}, N.~{Komodakis}, Dynamic edge-conditioned filters in
  convolutional neural networks on graphs, in: Conference on Computer Vision
  and Pattern Recognition (CVPR), 2017, pp. 29--38.

\bibitem{sift}
M.~Brown, S.~S{\"{u}}sstrunk, {Multi-spectral SIFT for scene category
  recognition}, in: Conference on Computer Vision and Pattern Recognition
  (CVPR), 2011, pp. 177--184.

\bibitem{hog}
L.~Xie, F.~Lee, L.~Liu, Z.~Yin, Y.~Yan, W.~Wang, J.~Zhao, Q.~Chen, {Improved
  spatial pyramid matching for scene recognition}, Pattern Recognition 82
  (2018) 118--129.

\bibitem{places365}
B.~Zhou, A.~Lapedriza, A.~Khosla, A.~Oliva, A.~Torralba, {Places: A 10 million
  Image Database for Scene Recognition}, IEEE Transactions on Pattern Analysis
  and Machine Intelligence (2017).

\bibitem{vgg}
K.~Simonyan, A.~Zisserman, {Very Deep Convolutional Networks for large-scale
  Image Recognition}, International Conference on Learning Representations
  (2015).

\bibitem{resnet}
K.~He, X.~Zhang, S.~Ren, J.~Sun, {Deep Residual Learning for Image
  Recognition}, in: Conference on Computer Vision and Pattern Recognition
  (CVPR), IEEE, 2016, pp. 770--778.

\bibitem{semantincSceneRecognition}
M.~George, M.~Dixit, G.~Zogg, N.~Vasconcelos, {Semantic Clustering for Robust
  Fine-Grained Scene Recognition}, in: European Conference on Computer Vision
  (ECCV), Springer International Publishing, 2016, pp. 783--798.

\bibitem{chengraphscenelayout}
G.~{Chen}, X.~{Song}, H.~{Zeng}, S.~{Jiang}, Scene recognition with
  prototype-agnostic scene layout, IEEE Transactions on Image Processing 29
  (2020) 5877--5888.

\bibitem{xiescenesurvey}
L.~Xie, F.~Lee, L.~Liu, K.~Kotani, Q.~Chen, Scene recognition: A comprehensive
  survey, Pattern Recognition 102 (2020) 107205.

\bibitem{multimodalfusion2016}
H.~{Zhu}, J.~{Weibel}, S.~{Lu}, Discriminative multi-modal feature fusion for
  rgbd indoor scene recognition, in: 2016 IEEE Conference on Computer Vision
  and Pattern Recognition (CVPR), 2016, pp. 2969--2976.

\bibitem{songlearningeffective}
X.~{Song}, S.~{Jiang}, L.~{Herranz}, C.~{Chen}, Learning effective rgb-d
  representations for scene recognition, IEEE Transactions on Image Processing
  28~(2) (2019) 980--993.

\bibitem{dfnet}
Y.~Li, J.~Zhang, Y.~Cheng, K.~Huang, T.~Tan, Df2net: Discriminative feature
  learning and fusion network for rgb-d indoor scene classification, in: AAAI,
  2018.

\bibitem{rgbdscenecategorizationmultimodal}
Z.~Cai, L.~Shao, {RGB-D Scene Classification via Multi-modal Feature Learning},
  Cognitive Computation (2018).

\bibitem{mapnet}
Y.~Li, Z.~Zhang, Y.~Cheng, L.~Wang, T.~Tan, {MAPNet: Multi-modal attentive
  pooling network for RGB-D indoor scene classification}, Pattern Recognition
  90 (2019) 436--449.

\bibitem{songobjecto-to-object}
X.~{Song}, S.~{Jiang}, B.~{Wang}, C.~{Chen}, G.~{Chen}, Image representations
  with spatial object-to-object relations for rgb-d scene recognition, IEEE
  Transactions on Image Processing 29 (2020) 525--537.

\bibitem{trecnet}
D.~Du, L.~Wang, H.~Wang, K.~Zhao, G.~Wu, {Translate-to-Recognize Networks for
  RGB-D Scene Recognition}, in: Conference on Computer Vision and Pattern
  Recognition (CVPR), IEEE, 2019, pp. 11828--11837.

\bibitem{ragc}
A.~Mosella-Montoro, J.~Ruiz-Hidalgo, {Residual Attention Graph Convolutional
  Network for Geometric 3D Scene Classification}, in: International Conference
  on Computer Vision Workshop (ICCVW), IEEE, 2019, pp. 4123--4132.

\bibitem{MultiViewSu3DShapeRecog}
H.~Su, S.~Maji, E.~Kalogerakis, E.~Learned-Miller, {Multi-view Convolutional
  Neural Networks for 3D Shape Recognition}, in: 2015 IEEE International
  Conference on Computer Vision (ICCV), IEEE, 2015, pp. 945--953.

\bibitem{SnapNet}
A.~Boulch, B.~L. Saux, N.~Audebert, {Unstructured Point Cloud Semantic Labeling
  Using Deep Segmentation Networks}, in: I.~Pratikakis, F.~Dupont,
  M.~Ovsjanikov (Eds.), Eurographics Workshop on 3D Object Retrieval, The
  Eurographics Association, 2017.

\bibitem{SnapNetR}
J.~Guerry, A.~Boulch, B.~L. Saux, J.~Moras, A.~Plyer, D.~Filliat, {SnapNet-R:
  Consistent 3D Multi-view Semantic Labeling for Robotics}, in: 2017 IEEE
  International Conference on Computer Vision Workshops (ICCVW), IEEE, 2017,
  pp. 669--678.

\bibitem{dai3DMV}
A.~Dai, M.~Nie{\ss}ner, {3DMV: Joint 3D-Multi-view Prediction for 3D Semantic
  Scene Segmentation}, in: V.~Ferrari, M.~Hebert, C.~Sminchisescu, Y.~Weiss
  (Eds.), Computer Vision -- ECCV 2018, Springer International Publishing,
  Cham, 2018, pp. 458--474.

\bibitem{VoxNet}
D.~Maturana, S.~Scherer, {VoxNet: A 3D Convolutional Neural Network for
  real-time object recognition}, in: IEEE International Conference on
  Intelligent Robots and Systems, Vol. 2015-Decem, 2015, pp. 922--928.

\bibitem{3DShapeNets}
Z.~Wu, S.~Song, A.~Khosla, F.~Yu, L.~Zhang, X.~Tang, J.~Xiao, {3D ShapeNets: A
  deep representation for volumetric shapes}, in: Proceedings of the IEEE
  Computer Society Conference on Computer Vision and Pattern Recognition, Vol.
  07-12-June, IEEE, 2015, pp. 1912--1920.

\bibitem{VolumetricMultiViewCNN}
C.~R. Qi, H.~Su, M.~Niessner, A.~Dai, M.~Yan, L.~J. Guibas, {Volumetric and
  Multi-View CNNs for Object Classification on 3D Data}, 2016 IEEE Conference
  on Computer Vision and Pattern Recognition (CVPR) (2016) 5648--5656.

\bibitem{ScanNet}
A.~Dai, A.~X. Chang, M.~Savva, M.~Halber, T.~Funkhouser, M.~Nie{\ss}ner,
  {ScanNet: Richly-annotated 3D reconstructions of indoor scenes}, in:
  Proceedings - 30th IEEE Conference on Computer Vision and Pattern
  Recognition, CVPR 2017, Vol. 2017-Janua, IEEE, 2017, pp. 2432--2443.

\bibitem{SEGCloud}
L.~Tchapmi, C.~Choy, I.~Armeni, J.~Gwak, S.~Savarese, {SEGCloud: Semantic
  segmentation of 3D point clouds}, in: Proceedings - 2017 International
  Conference on 3D Vision, 3DV 2017, 2017, pp. 537--547.

\bibitem{OctreeGeneratingNetworks}
M.~Tatarchenko, A.~Dosovitskiy, T.~Brox, {Octree Generating Networks: Efficient
  Convolutional Architectures for High-resolution 3D Outputs}, in: Proceedings
  of the IEEE International Conference on Computer Vision, Vol. 2017-Octob,
  IEEE, 2017, pp. 2107--2115.

\bibitem{dgcnn}
Y.~Wang, Y.~Sun, Z.~Liu, S.~E. Sarma, M.~M. Bronstein, J.~M. Solomon, {Dynamic
  Graph CNN for Learning on Point Clouds}, ACM Transactions on Graphics (TOG)
  (2019).

\bibitem{ScarselliGraphNeuralNetwork}
F.~Scarselli, M.~Gori, {Ah Chung Tsoi}, M.~Hagenbuchner, G.~Monfardini, {The
  Graph Neural Network Model}, IEEE Transactions on Neural Networks 20~(1)
  (2009) 61--80.

\bibitem{Qui3DGNNSS}
X.~Qi, R.~Liao, J.~Jia, S.~Fidler, R.~Urtasun, {3D Graph Neural Networks for
  RGBD Semantic Segmentation}, in: 2017 IEEE International Conference on
  Computer Vision (ICCV), IEEE, 2017, pp. 5209--5218.

\bibitem{kipf2017semi}
T.~N. Kipf, M.~Welling, {Semi-Supervised Classification with Graph
  Convolutional Networks}, in: International Conference on Learning
  Representations (ICLR), 2017.

\bibitem{FeaStNetCVPR2018}
N.~Verma, E.~Boyer, J.~Verbeek, {FeaStNet: Feature-Steered Graph Convolutions
  for 3D Shape Analysis}, in: CVPR 2018, 2018.

\bibitem{fusionet}
V.~Hegde, R.~Zadeh, Fusionnet: 3d object classification using multiple data
  representations, arXiv preprint arXiv:1607.05695 (2016).

\bibitem{splatnet}
H.~Su, V.~Jampani, D.~Sun, S.~Maji, E.~Kalogerakis, M.-H. Yang, J.~Kautz,
  Splatnet: Sparse lattice networks for point cloud processing, in: Proceedings
  of the IEEE conference on computer vision and pattern recognition, 2018, pp.
  2530--2539.

\bibitem{bcl}
V.~Jampani, M.~Kiefel, P.~V. Gehler, Learning sparse high dimensional filters:
  Image filtering, dense crfs and bilateral neural networks, in: Proceedings of
  the IEEE Conference on Computer Vision and Pattern Recognition, 2016, pp.
  4452--4461.

\bibitem{pointnet++}
C.~R. Qi, L.~Yi, H.~Su, L.~J. Guibas, Pointnet++: Deep hierarchical feature
  learning on point sets in a metric space, arXiv preprint arXiv:1706.02413
  (2017).

\bibitem{mvpnet}
M.~Jaritz, J.~Gu, H.~Su, Multi-view pointnet for 3d scene understanding, in:
  Proceedings of the IEEE/CVF International Conference on Computer Vision
  Workshops, 2019, pp. 0--0.

\bibitem{hha}
S.~Gupta, R.~Girshick, P.~Arbel{\'{a}}ez, J.~Malik, {Learning Rich Features
  from RGB-D Images for Object Detection and Segmentation}, in: European
  Conference on Computer Vision (ECCV), Springer International Publishing,
  2014, pp. 345--360.

\bibitem{Dynamicfilter}
X.~Jia, B.~De~Brabandere, T.~Tuytelaars, L.~V. Gool, {Dynamic Filter Networks},
  in: D.~D. Lee, M.~Sugiyama, U.~V. Luxburg, I.~Guyon, R.~Garnett (Eds.),
  Advances in Neural Information Processing Systems 29, Curran Associates,
  Inc., 2016, pp. 667--675.

\bibitem{SUNRGBD}
S.~{Song}, S.~P. {Lichtenberg}, J.~{Xiao}, Sun rgb-d: A rgb-d scene
  understanding benchmark suite, in: 2015 IEEE Conference on Computer Vision
  and Pattern Recognition (CVPR), 2015, pp. 567--576.

\bibitem{NYUV2}
P.~K. Nathan~Silberman, Derek~Hoiem, R.~Fergus, Indoor segmentation and support
  inference from rgbd images, in: ECCV, 2012.

\bibitem{pytorch}
A.~Paszke, S.~Gross, S.~Chintala, G.~Chanan, E.~Yang, Z.~DeVito, Z.~Lin,
  A.~Desmaison, L.~Antiga, A.~Lerer, Automatic differentiation in pytorch, in:
  NIPS-W, 2017.

\bibitem{pygeometric}
M.~Fey, J.~E. Lenssen, Fast graph representation learning with {PyTorch
  Geometric}, in: ICLR Workshop on Representation Learning on Graphs and
  Manifolds, 2019.

\bibitem{nsamples}
Y.~{Cui}, M.~{Jia}, T.~{Lin}, Y.~{Song}, S.~{Belongie}, Class-balanced loss
  based on effective number of samples, in: 2019 IEEE/CVF Conference on
  Computer Vision and Pattern Recognition (CVPR), 2019, pp. 9260--9269.

\bibitem{radam}
L.~Liu, H.~Jiang, P.~He, W.~Chen, X.~Liu, J.~Gao, J.~Han, On the variance of
  the adaptive learning rate and beyond, in: International Conference on
  Learning Representations, 2020.

\end{thebibliography}

\end{document}